\documentclass[10pt,
    twocolumn,
    ]{article}

\usepackage{iccv}
\usepackage{times}
\usepackage{epsfig}
\usepackage{graphicx}
\usepackage{amsmath}
\usepackage{amssymb}

\usepackage[pagebackref=true,breaklinks=true,
    colorlinks,bookmarks=false]{hyperref}

\iccvfinalcopy 

\usepackage{multirow}
\usepackage{pbox}
\usepackage{float}
\usepackage{listings}
\usepackage{caption}
\usepackage{verbatim}
\usepackage{graphicx}
\usepackage{subfig}
\usepackage{overpic}

\begin{document}

\title{Deformably-Scaled Transposed Convolution}

\author{
Stefano B. Blumberg, \
Daniele Rav\'{i}, \
Mou-Cheng Xu, \\
Matteo Figini, \
Iasonas Kokkinos, \
Daniel C. Alexander
\\
University College London (UCL)
}

\maketitle

\begin{abstract}
\noindent Transposed convolution is crucial for generating high-resolution outputs, yet has received little attention compared to convolution layers.   
In this work we revisit transposed convolution and introduce a novel layer that allows us to place information in the image selectively and choose the `stroke breadth' at which the image is synthesized, whilst incurring a small additional parameter cost. 
For this we introduce three ideas: firstly, we regress offsets to the positions where the transpose convolution results are placed; secondly we broadcast the offset weight locations over a learnable neighborhood; and thirdly we use a compact parametrization to share weights and restrict offsets.  
We show that simply substituting upsampling operators with our novel layer produces substantial improvements across  tasks as diverse as instance segmentation, object detection, semantic segmentation, generative image modeling, and 3D magnetic resonance image enhancement, while outperforming all existing variants of transposed convolutions.  Our novel layer can be used as a drop-in replacement for 2D and 3D upsampling operators and the code will be publicly available.
\end{abstract}

\begin{figure}[ht]
\centering
    \includegraphics[width=0.7\linewidth]{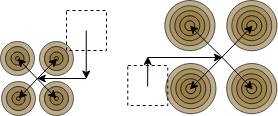}
    \caption{Deformably-Scaled Transposed Convolution (DSTC)  modifies the ``transmitive field'' of a neuron, allowing us to place information in the output layer in a more flexible manner than standard Transposed Convolution: instead of associating the input neuron's position (shown in a dashed box) to its adjacent output positions (a $2 \times 2$ grid of positions), we introduce a displacement vector, followed by a continuous dilation, that places the $2\times 2$ grid to a controllable, flexible set of positions; we further introduce a controllable kernel width, that allows us to set a `stroke width' that can accommodate for instance the gaps caused in the grid by the dilation factor. 
}
\end{figure}

\section{Introduction}

The convolution operations used in Convolutional Neural Networks (CNNs) have been recently modified to control feature acuity \cite{dai2017,zhu2019}, scale-invariance \cite{qin2018}, translation-invariance \cite{zhang2019}, or context-awareness \cite{chen2016,chen2017}, providing us with a rich arsenal of tools to improve image encoding. This is not the case for image decoding, where most architectures choose between three options: i) nearest-neighbors interpolation e.g. in \cite{karras2018,park2019,osendorfer2014}, ii) bilinear interpolation e.g. in \cite{dong2016,chen2018,zha2017,zhao2018,yu2017}, iii) transposed convolution (TC), also known as deconvolution, or fractionally-strided convolution, used e.g. in \cite{zeiler2011,isola2017,radford2016,ronneberger2015,blumberg2018,he2017,ravi2022}. We hypothesize that substantial improvements in decoding-based tasks can be achieved by better designing the decoding counterparts to advanced encoding layers.
\\ 
\indent In particular we can attribute the success of deformable convolutions \cite{dai2017,zhu2019} to the treatment of scale as a nuisance parameter that is first estimated and then used to deliver invariance; and we can understand smoothing-based downsampling \cite{zhang2019} as a remedy to the aliasing incurred by naive image decimation. But the same problems plague decoding, where one may need to create an output at multiple scales or under non-rigid deformations, while checkerboard artifacts can occur \cite{odena2016} when naively transmitting features.
\\
\indent Motivated by this observation, we introduce a new upsampling layer in deep learning: the Deformably-Scaled Transposed Convolution (DSTC) that leverages concepts from deformable convolution \cite{dai2017,zhu2019} and aliasing-free downsampling \cite{oppenheim1999,zhang2019} in order to exert stronger control on the image decoding task. 
The deformable aspect of our layer comes from modifying the fixed displacement pattern used for TC by learnable offsets. This allows an input neuron to transmit its signal to a learnable neighborhood that can be adaptively scaled or deformed non-rigidly.  Changing the ``transmitive field'' of a neuron during the TC operation can however have undesirable effects on the output, documented e.g. in \cite{odena2016} for CNNs, or more easily understood as interpolation distortion in the linear case \cite{oppenheim1999}.  The DSTC mitigates this, by removing high-frequency artefacts through a learnable interpolation kernel.  As such, the DSTC uses two additional modules than the TC, requiring two separate heads added to the input feature map of our operation.  
\\
\indent Furthermore, based on the hypothesis that the DSTC has an unnecessary number of degrees of freedom, we use a parametrization.  We parametrize the offsets by restricting the input-output location mapping to a simple change in location for each dimension and a change in scale; that consists of learning the dilation factor, and parametrize the interpolation kernel with weight sharing.  Thus, our DSTC requires only a small increase of parameters over the TC.
\\
\indent We evaluate the DSTC, showing its general purpose nature, via simple substitutions of upsampling layers across a diverse set of tasks, without changing network architecture and without modifying the training procedure.  Our tasks are: object detection and instance segmentation with COCO using the feature pyramids networks \cite{lin2017} with the the Mask R-CNN \cite{he2017}, semantic segmentation on VOC \cite{everingham2015} using the HRNet \cite{sun2019}, generating scaled CelebA faces \cite{CelebA} with the DCGAN \cite{radford2016},  3D diffusion magnetic resonance image (MRI) enhancement on human brains \cite{sotiropoulos2013} with the Deeper Image Quality Transfer (DIQT) network \cite{blumberg2018} -- where we obtain state-of-the-art results.
\\
\indent We demonstrate that the DSTC produces improved results in the 2D tasks, compared to the standard TC and other commonly-used upsampling operators such as nearest-neighbors interpolation and bilinear interpolation.  The DSTC also outperforms more recent upsampling operators: the Pixel-Wise Shuffle (a.k.a. Sub-pixel Convolutional Layer) \cite{shi2016}, the Transposed Pixel-Adaptive Convolution \cite{su2019}, and Content-Aware ReAssembly of FEatures (CARAFE) \cite{wang2019}.  In addition in the 2D experiments we show the benefits of adding our two modules along with the benefits of the parametrization.  The code will be publicly available.

\section{Related Work}s

\noindent \textbf{Adaptive Convolutions} The first approach to spatially adapt features in deep learning was the Spatial Transformer Networks \cite{jaderberg2015}, which learnt how to effectively warp the entire input feature map.  A more effective approach was the Deformable Convolutional Networks (DCNs) \cite{dai2017,zhu2019}, which modified the sampling locations of a convolutional layer, where the values to augment the sampling locations were the output of an additional convolutional layer.  The Active Convolutional Unit \cite{jeon2017}, inspired by synapses, proposed a generalization of the convolutional operator, which may have different forms of receptive fields and takes in fractional pixel coordinates.  We reformulate concepts from DCNs into our upsampling paradigm.
\\
\\ \textbf{Upsampling Operators in Deep Learning} The most commonly used upsampling operators in computer vision are nearest-neighbor interpolation and bilinear interpolation which have no trainable parameters, are lightweight, are computationally inexpensive, and use strictly local information.  The TC (a.k.a deconvolution, fractionally-strided convolution) \cite{zeiler2011}, is the most commonly-used upsampling layer in deep learning for computer vision that has trainable parameters.  Here, individual pixels in the low-resolution input image are iteratively ''convolved" with a spatially-invariant weights/filters and the output is summed over target locations in the high-resolution space.  The relationship between the input and target locations, is the inverse of the ubiquitous convolutional layer.  A good overview of classical upsampling operators is \cite{wojna2018}.  More recent upsampling operators include the Sub-pixel Convolutional Layer \cite{shi2016}, which is a pixel-wise shuffle, and the Context-Aware ReAssembly of FEatures (CARAFE) \cite{wang2019}, which aggregates contextual features and generates adaptive kernels during training.  We enrich the modelling capacity of the TC, by integrating our two modules into its operation.
\\
\\ \textbf{Anti-Aliasing in Deep Learning} Avoiding artefacts and distortions caused by aliasing is a classical problem in signal processing \cite{oppenheim1999}.  Recently \cite{zhang2019} addressed anti-aliasing in the context of deep learning, by using simple spatial blurs before downsampling operations, to both improve network performance and improve robustness to shift-based adversarial attacks.  Furthermore, \cite{zou2020} extended learnt a low-pass filtering layer that adapts to various frequencies in images, to avoid aliasing.  We use concepts from anti-aliasing in the DSTC, by learning a linear combination of Gaussian kernels, which is used to interpolate regressors (in our case the target location of the TC operation) in the target feature space.

\begin{table*}[ht]
\centering
\resizebox{\linewidth}{!}{
\begin{tabular}{c c c c  }
\multicolumn{1}{c}{\multirow{2}{*}{\centering Layer }} &
\multicolumn{3}{c}{\multirow{1}{*}{\centering Parameters}} \\
\cline{2-4}
 & Convolutional Weights $ W $ & Offsets $conv_{\Delta p} $ & Interplation Kernel $ conv_{\Sigma} $ \\
\hline
Transposed Convolution (TC) & $ K^{D}C_{i}C_{o} + C_{o} $ & -- & -- \\
\hline
Deformably-Scaled Transposed Convolution (DSTC) non-parametrized w. bilinear interpolation kernel & $ K^{D}C_{i}C_{o} + C_{o} $ &  $ 3^{D}C_i \cdot DK^{D} $ & -- \\
Deformably-Scaled Transposed Convolution (DSTC) non-parametrized & $ K^{D}C_{i}C_{o} + C_{o} $  & $ 3^{D}C_i \cdot DK^{D} $ & $ 3^{D}C_i \cdot sDK^{D} $ \\ 
Deformably-Scaled Transposed Convolution (DSTC) parametrized  & $ K^{D}C_{i}C_{o} + C_{o} $ & $ 3^{D}C_{i} \cdot (D+1) $ & $ 3^{D}C_i \cdot s $ \\ 
\end{tabular}
}
\caption{Number of parameters in our DSTC layers, transposed convolutional weight kernel size $ K $, input/output channels $C_{i}/C_{o}$, spatial dimension $ D =2,3 $, interpolation kernel has $ s $ Gaussian variances.}\label{module_params:table}
\end{table*}

\section{Methods}

\noindent In this section we introduce: i) learning offsets for the TC, ii) a learnt interpolation kernel for these offsets, iii) a parametrization for (i),(ii).

\noindent 
\\ \textbf{Notation} Suppose we have an input feature map $ X \in \mathbb{R}^{C_i \times H_i \cdot W_i (\cdot D_i)} $ and target feature map $ Y \in \mathbb{R}^{C_o \times H_o \cdot W_o (\cdot D_o)}$ where $ (H_i,W_i,D_i) \leq (H_o,W_o,D_o) $,  which may be of spatial dimension $ D = 2,3 $.  The values $ C_{i/o} $ is the number of channels and $ H_{i/o},W_{i/o} (,D_{i/o}) $ is the height, width (, depth) of the feature maps.
\\ 
\\ \textbf{Preliminaries} To better explain a TC from $ X $ to $ Y $ we first consider the related (standard) convolution from $ Y $ to $ X $.  Given a location $ p_0 $ in $ X $, its value $ X(p_0) $ depends on first sampling on a grid $ \mathcal{R} $ on $ Y $, then summing the samples weighted by a weight $ W $.   In 2D, the reference grid that corresponds to kernel shape of identical height and width equal to $ K $, with dilation $ 1 $ is
\begin{equation}
\mathcal{R} = \{(i,j) \in \mathbb{Z}^{2} \ \ - \lfloor K/2 \rfloor \leq i,j \leq  \lfloor \frac{K-1}{2} \rfloor  \}.
\end{equation}
Then for each location $ p_0 $ on $ X $, the standard convolution is the linear operation
\begin{equation}\label{TCEQ}
\begin{gathered}
X(p_0) = \sum_{p_n} Y(r(p_0) + p_n) \cdot W(p_n) \\
\end{gathered}
\end{equation}
where $ n = 1...|\mathcal{R}| $ enumerates the locations $ p_{n} $, in $ \mathcal{R} $ and $ r $ maps locations in $ Y $ to locations in $ X $, to take into account possible changes of resolution.  The locations $ \{(r(p_0) + p_n) \} $ in $ Y $ is called the receptive field of the pixel $ X(p_0) $.  
\\
\indent Now the related TC (a.k.a. strided convolution, deconvolution) of the above operation, is a linear operation from $ X $ to $ Y $.  With the same notation as before, the TC is defined for each location $ p_0 $ in $ X $ as
\begin{equation}\label{TransIndiv}
\begin{gathered}
Y(r(p_0) + p_n) = X(p_0) \cdot W(p_n) \ \ \ n=1...,|\mathcal{R}| \\
W \in \mathbb{R}^{C_{i} \times C_{o} \times K_{1} \times K_{2} (\times K_{3})}
\end{gathered}
\end{equation}
iterated over the locations $ p_{0}^{l} \ l=1...H_{i}\cdot W_{i}(\cdot D_{i}) $ in $ X $ and sum the outputs, to obtain the value of location $ \tilde{p}_0 $ in $ Y $:
\begin{equation}
Y(\tilde{p}_0) = \sum_{i,p_n} X(p_{0}^{l}) \cdot W(p_n) \mathbb{I}_{\tilde{p}_0 =  r(p_{0}^{l}) + p_n}.
\end{equation}

\begin{figure*}[ht]
\begin{center}
   \begin{overpic}
         [width=\linewidth]{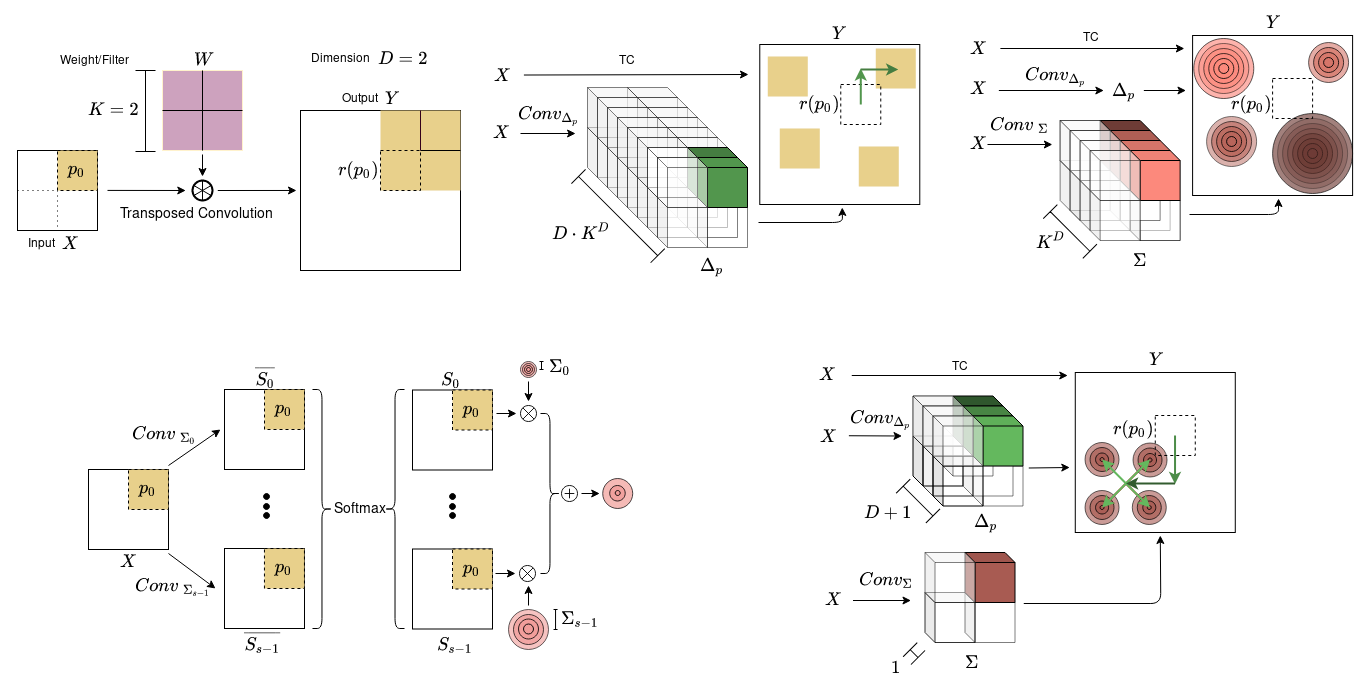}
         \put(6, 27){\footnotesize Transposed Convolution (Baseline)}
         \put(35, 27){\footnotesize DSTC (non-parametrized w. bilinear interpolation kernel)}
         \put(78, 27){\footnotesize DSTC (non-parametrized)}
         \put(15, -0.5){\footnotesize Our Learnt Interpolation Kernel}
         \put(67.5, -0.5){\footnotesize DSTC (parametrized)}
   \end{overpic}
\end{center}
\vspace*{-2.5mm}
\captionof{figure}{An illustration of the Deformably-Scaled Transposed Convolution (DSTC) layers, number of parameters in table~\ref{module_params:table}.
The DSTC has two additional modules to the baseline transposed convolution to learn a tensor $\Delta p $, corresponding to the offsets of the target locations and tensor $ \Sigma $, corresponding to the spread of the interpolation kernel for the target locations.  The parametrization uses weight sharing and restricts the offsets to a simple shift in location and scale, which corresponds to learning the dilation.
}\label{fig:TC}\label{fig:Learning_The_Offsets}\label{fig:Learning_An_Interpolation_Kernel}\label{fig:Scoring_For_Fixed_Gaussians}\label{fig:DSTC_Parametrization}
\end{figure*}
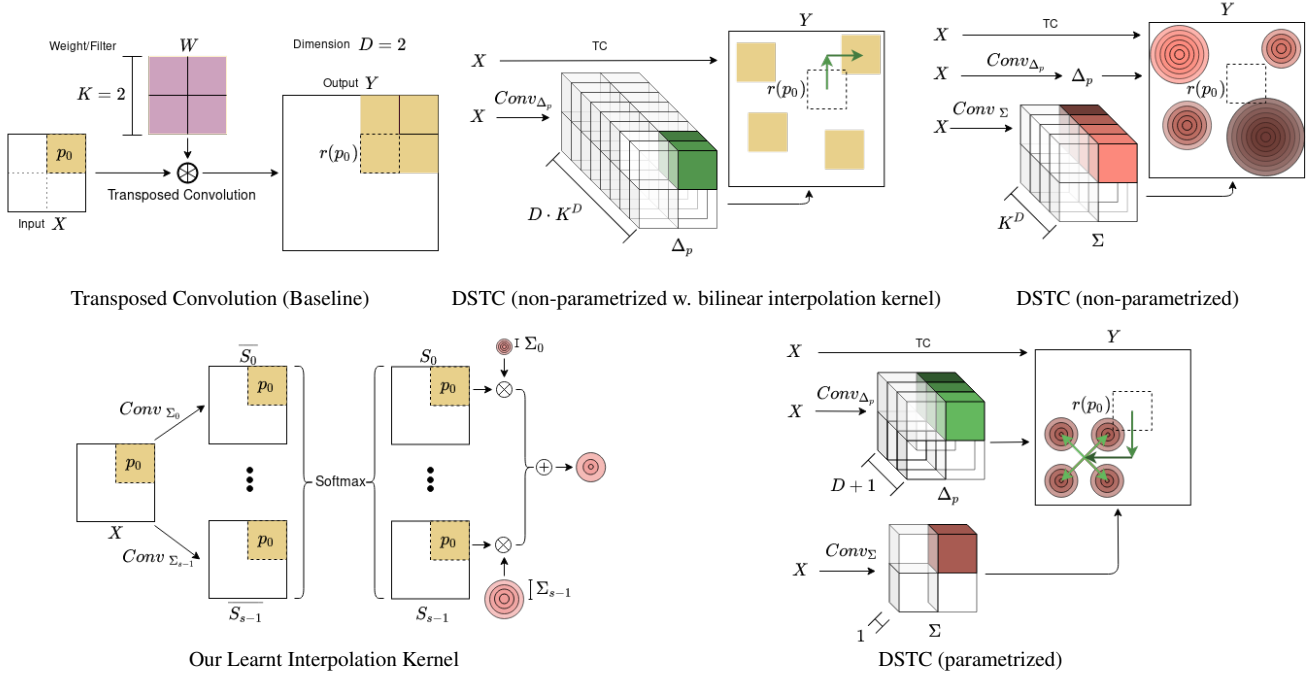

\noindent We provide an illustration of the TC operation in figure \ref{fig:TC}.
\\
\\ \textbf{Learning the Target Offsets} TCs are restricted by the fixed relationships between the input and target locations, which limits the modelling capacity and may produce artefacts. Instead, learning the offsets is a better balance between the strong convolutional prior and the efficiency to learn potentially useful data-informed features.  We reformulate the approach in \cite{dai2017} and learn the offsets for the target locations of the TC, via a 3x3(x3) convolution of $ X $
\begin{equation}\label{OffConv}
\Delta p = conv_{\Delta p}(X) \ \ \ \ \Delta p \in \mathbb{R}^{ D \cdot |\mathcal{R}| \times H_i \cdot W_i (\cdot D_i)}.
\end{equation}
where the value $ \Delta p (d \cdot n, l) $  is the offset for weight/sample index $ n $, input location $ p^{l}_0, \ l = 1...H_i \cdot W_i (\cdot D_i) $ in spatial dimension $ d $.  We denote the offset locations as 
\begin{equation}\label{OffsetEq}
q(n,l) = r(p_{0}^{l}) + p_n + (\Delta p(1 \cdot n, p_{0}^{l})... \Delta p(D \cdot n, p_{0}^{l}))
\end{equation}
and we replace equation-\ref{TransIndiv} with
\begin{equation}\label{TransOffIndiv}
Y(q(n,l)) = X(p_{0}^{l}) \cdot W(p_n)
\end{equation}
which we illustrate 
in figure \ref{fig:Learning_The_Offsets}.
\\
\\ \noindent \textbf{Learning Interpolation Kernels for the Offsets} As the offset locations are usually not integers, we need to interpolate these fractional positions to integer positions $ p $ in $ Y $ and sum over target locations

\begin{equation}\label{InterpEq}
\begin{gathered}
Y(p) = \sum_{n,l} Y^{n,l}(p) \\ 
Y^{n,l}(p) = \sum_q G^{n,l}(q(n,l),p) \cdot Y(q(n,l))
\end{gathered}
\end{equation}
with an interpolation kernel $ G^{n,l} $, of size $ K_{\Sigma} > 0 $, which may differ depending on location $ p_{0}^{l} $ in $ X $ and weight index $ n $.  The most commonly used interpolation kernel (e.g. in \cite{dai2017,zhu2019,jaderberg2015}) is the bilinear/trilinear kernel
\begin{equation}
\begin{gathered}
G(q,p) := g(q_x,p_x)\cdot g(q_y,p_y)(\cdot g(q_z,p_z)) \\ 
g(a,b):= max(0, 1-|a-b|)
\end{gathered}
\end{equation}
which does not depend on the weight index or input location i.e. $ G^{n,l}(q,p) = G(q,p) $.  It has no trainable parameters and is of size $ K_{\Sigma} = 2 $.  To enhance the modelling capacity to handle deformations, we propose to learn $ G^{n,l}(q,p) $ in a dense fashion, i.e. for different $ n=1...|\mathcal{R}|, \ l=1...H_i \cdot W_i (\cdot D_i) $.  We will also increase $ K_{\Sigma} $, which increases the receptive field of pixels in $ Y $. 
\\ \indent We propose that the layer learn a multi-scale smoother, for each regressor (depending on different $ n,l $ ) in the target feature map.  We propose a scoring system for $ s $ Gaussian interpolation kernels, which are fixed a priori.  First we choose hyperparameters $ 0 < \Sigma_{0} < ... < \Sigma_{s-1} $, the variances of $ s $ Gaussian blurs.  Then we use a 3x3(x3) convolution from the input feature map, to learn $ s $ scoring maps
\\
\begin{equation}\label{ScoreConv}
\begin{gathered}
{[}\overline{S}_{1}...\overline{S}_{s-1}{]} = \overline{S} = conv_{\Sigma}(X) \ \ \ \ j=0...s-1
\\ \overline{S}_j(n,l) \in \mathbb{R}^{|\mathcal{R}| \times H_i \cdot W_i (\cdot D_i)},
\end{gathered}
\end{equation}
where $ \overline{S}_{j} $ is normalized with a sigmoid if $ s = 1 $, or a softmax if $ s \geq 2 $.  For fixed offset $ q(n,l) $ in equation \ref{OffsetEq}, we express the interpolation kernel from equation \ref{InterpEq} as
\begin{equation}\label{GaussInt}
\begin{gathered}
\begin{split}
G^{n,l}&(p,q(n,l)) = 
\\ & \sum_{j=0}^{s-1} N(j) S_{j}(n,l)e ^{ - \frac{||p-q(n,l)||_{2}^{2}}{2\Sigma_j}}\mathbb{I}_{ \{||p-q(n,l)||_\infty < K_\Sigma \} }
\end{split}
\end{gathered}
\end{equation}
where $ N(j)$ is a normalization term.

\begin{figure*}[ht]
\begin{center}
   \includegraphics[width=0.32\linewidth]{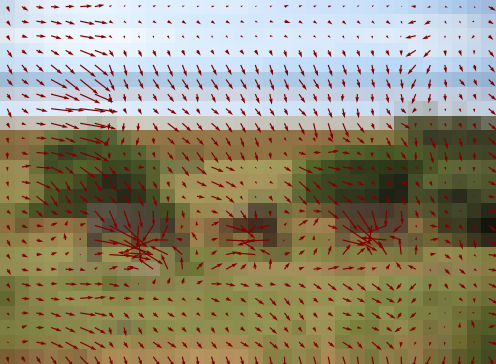}
   \includegraphics[width=0.32\linewidth]{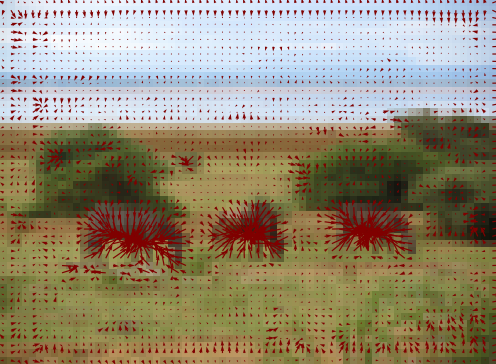}
   \includegraphics[width=0.32\linewidth]{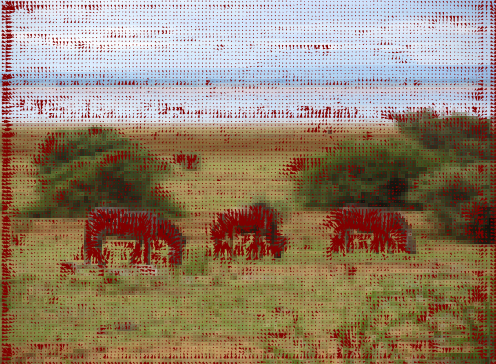}
   \\
   \includegraphics[width=0.32\linewidth]{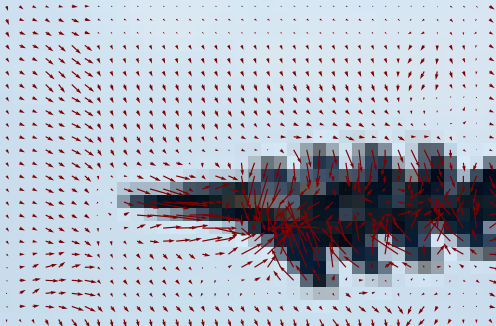}
   \includegraphics[width=0.32\linewidth]{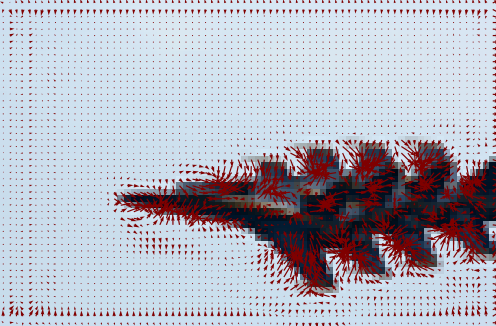}
   \includegraphics[width=0.32\linewidth]{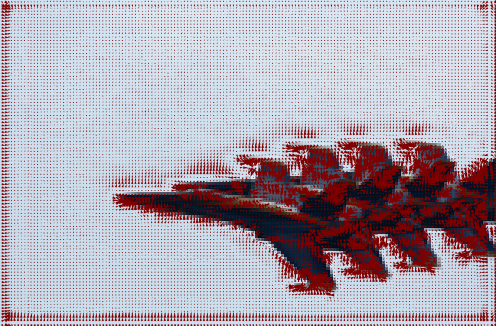}
\end{center}
   \caption{Learnt DSTC offsets $ \Delta p $ in the Feature Pyramid Network (FPN) on  (left-to-right) low,middle,high resolution features in the FPN.  The DSTC maps each object and its boundaries, towards its centre.  
   }\label{DSTC_offsets:fig}
\end{figure*}

We illustrate this module in figure~\ref{fig:Learning_An_Interpolation_Kernel}.
\\
\\ \noindent \textbf{A Parametrization for Learning the Offsets and Learning the Offset Interpolation Kernel} Our DSTC layer aims to be adaptable towards geometric changes of similar object representations, in different input feature maps.  However, our modules might have too many degrees of freedom as it might be unnecessary to have such a high-dimensional mapping.  Therefore we propose a parametrization for both the offsets and learnt interpolation kernels that promotes learning lower-dimensional manifolds, that have adequate modelling capacity. 
\\ \indent We parametrize the offsets by restricting their geometric shift to a simpler change in location and scale.  More specifically, given an input location $ p_{0}^{l} l =1...H_i \cdot W_i (\cdot D_i) $, we learn a spatial shift and the dilation factor for the offset locations $ q(1,l)...q(K,l) $ from equation \ref{OffsetEq}.  Instead of learning a tensor $ \Delta p \in \mathbb{R}^{ D \cdot |\mathcal{R}| \times H_i \cdot W_i (\cdot D_i)} $ in equation-\ref{OffConv}, we reduce the number of output channels in $ conv_{\Delta p} $ to learn $ \Delta p \in \mathbb{R}^{ (1+D) \times H_i \cdot W_i (\cdot D_i)} $.  The value $ \Delta p(1,l) $ is an expansion factor (the dilation of the TC) and the values $ \Delta p(1 + d) \ d=1...D $ correspond to the shift in target locations in spatial dimension $ d $.  
\\ \indent We also propose to parametrize our kernel learning approach via weight sharing, where we let $ G^{1,l} = ... = G^{|\mathcal{R}|,l} $ for different $ l $ in equation \ref{TransOffIndiv}.  Instead of learning tensors $ \overline{S}_{j} \in \mathbb{R}^{|\mathcal{R}| \times H_i \cdot W_i (\cdot D_i)} $, we learn a tensor $ \overline{S}_{j} \in \mathbb{R}^{1 \times H_i \cdot W_i (\cdot D_i)} $, by reducing the number of output channels in the convolutions $ conv_{\Sigma} $ in equation~\ref{ScoreConv}.
\\ \indent  We illustrate the parametrization in figure~\ref{fig:DSTC_Parametrization}, we note parametrizing our modules reduces both the number of computations and the number of parameters, as we lowered the output channels of $ conv_{\Delta p},conv_{\Sigma} $, see table~\ref{module_params:table}.
\\
\\ \textbf{Interpolation Kernel Hyperparameters} To set interpolation kernel size $ K_{\Sigma} $ and Gaussian variances $ \Sigma_{i} $ we analyzed Gaussian plots and conducted a brief hyperparameter search, presented in the supplementary materials.  We set $ K_{\Sigma} = 5 $.  When our layer is inserted in an intermediary upsampling layer of a network $ \Sigma_{0,1,2,3} =   2^{-2},2^{0},2^{2},2^{4}  $ and we initialize the dilation to 3.  When inserted in the last layer of a network we set $ \Sigma_{0,1,2,3} = \frac{1}{30},\frac{1}{2},1,2 $, to improve output image sharpness.  
\\
\\ \noindent \textbf{Implementation}  Our implementation is in Python with PyTorch \cite{PYTORCH} and is available in 2D or 3D.  During code development we used \cite{fey2019}.  The DSTC takes analogous arguments to the original TC and the user may choose  parametrized and nonparametrized versions of each module.  We illustrate code usage in the supplementary materials and the code will be publicly available.

\begin{table*}[ht]
\centering
\resizebox{\linewidth}{!}{
\begin{tabular}{c c c c c c c c c c c c c c c c}
\multicolumn{2}{c}{\centering Upsampling Operators} &
\multicolumn{1}{c}{\multirow{2}{*}{\pbox{1.5cm}{Params.}}} & 
\multicolumn{6}{c}{COCO Test-dev Box} & &
\multicolumn{6}{c}{COCO Test-dev Mask} \\
\cline{1-2} \cline{4-9} \cline{11-16}
FPN & Mask Head &  & $ AP $ & $ AP_{50} $  & $ AP_{75} $ & $ AP_{S} $ & $ AP_{M} $ & $ AP_{L} $ & & $ AP $ & $ AP_{50} $ & $ AP_{75} $ & $ AP_{S} $ & $ AP_{M} $ & $ AP_{L} $ \\ \hline
Nearest Neighbors & Transposed Conv. & 44.12M & $ 38.6 $ & $ 59.4 $  & $ 42.1 $ & $ 21.8 $ & $ 41.3 $ & $ 48.7 $ & & $ 35.0 $ & $ 56.4 $ & $ 37.4 $ & $ 18.6 $ & $ 37.3 $ & $ 46.0 $ \\ \hline 
\multicolumn{2}{c}{Transposed Conv.} & 46.22M & $ 38.3 $ & $ 58.9 $  & $ 41.7 $ & $ 21.6 $ & $ 40.8 $ & $ 48.5 $ & & $ 34.8 $ & $ 55.9 $ & $ 37.1 $ & $ 18.3 $ & $ 37.1 $ & $ 45.9 $ \\ 
\multicolumn{2}{c}{Nearest Neighbors + Conv.}  & 46.22M & $ 38.5 $ & $ 59.2 $  & $ 42.1 $ & $ 21.6 $ & $ 41.1 $ & $ 48.9 $ & & $ 35.1 $ & $ 56.3 $ & $ 37.6 $ & $ 18.4 $ & $ 37.4 $ & $ 46.2 $ \\ 
\multicolumn{2}{c}{Bilinear + Conv.}  & 46.22M & $ 38.4 $ & $ 59.2 $  & $ 41.9 $ & $ 21.8 $ & $ 40.9 $ & $ 48.6 $ & & $ 35.1$ & $ 56.4 $ & $ 37.6 $ & $ 18.8 $ & $ 37.3 $ & $ 46.1 $ \\ 
\multicolumn{2}{c}{Pixel-wise Shuffle + Conv.}  & 44.45M & $ 38.2 $ & $ 59.0 $  & $ 41.5 $ & $ 21.3 $ & $ 41.0 $ & $ 48.3 $ & & $ 34.8 $ & $ 56.0 $ & $ 37.2 $ & $ 18.2 $ & $ 37.2 $ & $ 45.8 $ \\ 
\multicolumn{2}{c}{Transposed Pixel-Adaptive Conv.} & 44.15M & $ 38.6 $ & $ 59.9 $  & $ 42.0 $ & $ 22.0 $ & $ 41.2 $ & $ 49.1 $ & & $ 35.3 $ & $ 56.7 $ & $ 37.9 $ & $ 18.9 $ & $ 37.5 $ & $ 46.4 $ \\ %
\multicolumn{2}{c}{CARAFE} & 44.16M & $ 39.0 $ & $ 60.1 $  & $ 42.5 $ & $ 22.4 $ & $ 41.6 $ & $ 49.3 $ & & $ \textbf{35.8} $ & $ 57.3 $ & $ \textbf{38.3} $ & $ 19.2 $ & $ 38.0 $ & $ \textbf{46.9} $ \\ 
\hline
\multicolumn{2}{c}{DSTC non-parmetrized w . bilinear kernel}  & 44.17M & $ 38.9 $ & $ 60.0 $  & $ 42.2 $ & $ 22.2 $ & $ 41.4 $ & $ 49.4 $ & & $ 35.2 $ & $ 56.6 $ & $ 37.6 $ & $ 18.9 $ & $ 37.4 $ & $ 46.5 $ \\ 
\multicolumn{2}{c}{DSTC non-parametrized} & 44.23M & $ \textbf{39.2} $ & $ \textbf{60.5} $  & $ \textbf{42.7} $ & $ 22.5 $ & $ \textbf{42.0} $ & $ 49.2 $ & & $ \textbf{35.8} $ & $ 57.6 $ & $ \textbf{38.3} $ & $ 19.2 $ & $ \textbf{38.2} $ & $ 46.6 $ \\ 
\multicolumn{2}{c}{DSTC parametrized}  & 44.15M & $ \textbf{39.2} $ & $ \textbf{60.5} $  & $ 42.6 $ & $ \textbf{22.7} $ & $ 41.8 $ & $ \textbf{49.6} $ & & $ \textbf{35.8} $ & $ \textbf{57.7} $ & $ \textbf{38.3} $ & $ \textbf{19.5} $ & $ 38.1 $ & $ 46.8 $ \\ 
\end{tabular}
}
\caption{Object detection and instance segmentation on the mmdetection \cite{mmdetection} implementation of the Mask-RCNN \cite{he2017} with Feature-Pyramid Networks (FPNs) \cite{lin2017}, with ResNet-50 backbone, trained on COCO.  The original configuration has three nearest neighbors in the FPN and a transposed convolution in the mask head, which we replace.}\label{mrcnn_results:table}
\end{table*}

\section{Experiments and Results}  

\noindent We show how our novel layer can improve network performance, by simply substituting upsampling operators in networks with the DSTC.  We demonstrate how our layer is more powerful than commonly-used upsampling layers: i) the prototype transposed convolution; ii) the nereast-neighbors interpolation followed by a convolution, iii) bilinear interpolation followed by a convolution, furthermore we compare the DSTC with three more recent, but less-used, upsampling operators: iv) the Pixel-wise Shuffle \cite{shi2016} followed by a convolution, v) the Transposed Pixel-Adaptive Convolution \cite{su2019}, vi) the Content-Aware ReAssembly of FEatures (CARAFE) \cite{wang2019}.  We use the official implementation for these operators, for the Transposed Pixel-Adaptive Convolution we learn the guidance feature via a convolution and bilinear upsampling layer, and set guidance channels to $ 7 $ such that the layer has the same number of parameters as the DSTC, for the CARAFE we do not compress the channels if the input channels is less than $ 64 $ (value used in \cite{wang2019}).

\subsection{Object Detection and Instance Segmentation with Mask-RCNN with FPN}\label{mrcnn_fpn:sec}

\noindent In this section, we use the Mask-RCNN \cite{he2017} with Feature-Pyramid Networks (FPNs) \cite{lin2017} to perform object detection and instance segmentation.  The FPNs \cite{lin2017}, illustrated in the supplementary materials, is a top-down pathway with four feature maps connected via three consecutive nearest-neighbors interpolation upsampling operations.  We use a Pytorch port of the original code from mmdetection \cite{mmdetection}.
\\
\indent We used the COCO 2017 \cite{lin2014} of 118K training images, 5K validation images (used for model development), and 40K test-dev images, of "common objects".  We obtained COCO Test-dev2019 scores by uploading results to the server.  We used the standard 1x training from \cite{mmdetection}, described in the supplementary materials.
\\
\indent  In our experiment we replace the three nearest-neighbors interpolation upsampling operations in the FPN and the TC in the mask head, with upsampling layers of kernel size $ K=3 $ (exculding CARAFE).  To reduce parameters and computational complexity and to have a fair comparison with CARAFE,  the DSTC and Transposed Pixel-Adaptive Convolution had $ 64 $ in/out channels, where we added a $ 1 \times 1 $ convolution before and after the operation to compress and expand the channel dimension.  We report quantitative results in table~\ref{mrcnn_results:table} and qualitative results in figure~\ref{mrcnn_results:fig}, noting that by simply altering four layers in the Mask-RCNN, we are able to make substantial improvements over the TC and other commonly-used operators.  This includes a small improvement over CARAFE, even though the CARAFE was developed for FPN-like architectures.  We also show learnt DSTC tensors in figure~\ref{DSTC_offsets:fig}.

\subsection{Semantic Segmentation with HRNet}\label{hrnet:sec}

\begin{figure*}[ht]
\begin{center}
   \includegraphics[width=\linewidth]{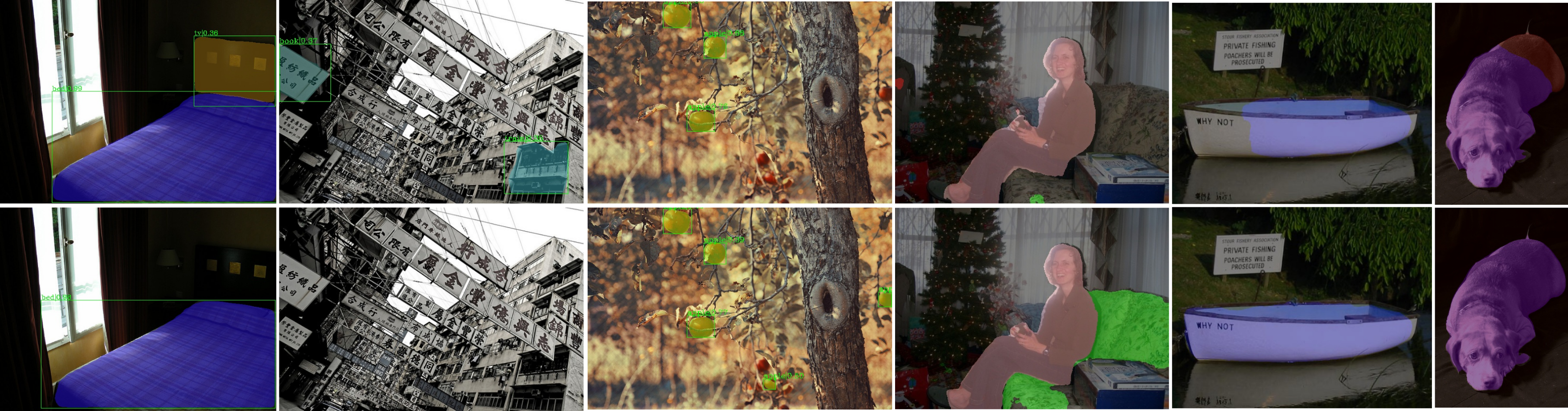}
\end{center}
   \caption{Visual comparison of results with transposed convolution (top) and DSTC (bottom).  First three images are instance segmentation results on COCO val 2017, second three images are semantic segmentation on VOC.}
\label{mrcnn_results:fig}\label{hrnet_results:fig}
\end{figure*}

\noindent The HRNet \cite{sun2019} has recently shown much promise across semantic segmentation, object detection and human pose estimation.  It has four stages, where each stage contains parallel branches of different resolution, at each successive stage, a lower-resolution branch is added.  At seven points in the HRNet, the feature maps at the different resolutions are fused onto all of the feature maps of different resolutions, combining representations at different scales.  Our task simply replaces the thirty-one bilinear upsampling operators, within the fusion layers (three layers upsample $ \times 8 $, ten layers upsample $ \times 4 $, eighteen layers upsample $ \times 2 $).  More specifically, we use the implementation from \cite{mmseg2020}, which uses a FCN head \cite{long2015} and the channel width multiplier of 48.  We train on the VOC 2012 augmented data set \cite{everingham2015} of 10582 images and our task is to classify the pixels in the 1449 VOC 2012 validation images into one of 21 classes.  We use the standard 20K schedule from \cite{mmseg2020}, rescaling the images to  $ 2048 \times 512 $, cropped to $ 512 \times 512 $, further details are in the supplementary materials.  Evaluation is performed at the single, original scale.
\begin{table}[ht]
\centering
\resizebox{\linewidth}{!}{
\begin{tabular}{c c c}
Upsampling Operators \ &  Params.  & VOC val mIOU \\
\hline
Bilinear  & 65.86M & 75.87 \\ 
\hline
Transposed Conv.      &   68.06M    & 76.17   \\ 
Nearest Neighbors + Conv.    &   68.06M   & 76.12  \\ 
Bilinear + Conv.     &   68.06M    & 76.02    \\ 
Pixel-wise Shuffle + Conv.      &  66.41M     &  75.42  \\ 
Transposed Pixel-Adaptive Conv. &  68.21M     &  75.92  \\ 
CARAFE      &   70.91M    & 75.94   \\ 
\hline
DSTC non-parametrized w. bil. kernel & 68.45M  & 76.43  \\ 
DSTC non-parametrized           & 69.23M & 76.38      \\ 
DSTC parametrized              & 68.21M    & \textbf{76.99} \\ 
\end{tabular}
}
\caption{Semantic segmentation on the mmsegmentation \cite{mmseg2020} implementation of the HRNet \cite{sun2019} width 48, with the FCN head \cite{long2015},trained with VOC 2012 Aug.  The original configuration has thirty-one bilinear interpolation upsampling layers, which we replace. }\label{hrnet:table}
\end{table}
\begin{table}[ht]
\centering
\resizebox{\linewidth}{!}{
\begin{tabular}{c c c}
Last Upsampling Op. in Generator \ &  Params.  &  FID \\
\hline
Transposed Conv.     &   6342K    & 29.6   \\
\hline
Nearest Neighbors + Conv.    &   6342K   & 36.1  \\ 
Bilinear + Conv.     &   6342K    & 85.3    \\ 
Transposed Pixel-Adaptive Conv. $ K = 3 $     &  6345K    &  32.7  \\ 
Transposed Pixel-Adaptive Conv. $ K = 5 $     &  6348K    &  31.5  \\ 
\hline
DSTC non-parametrized w. bil. interp. kernel & 6360K  & 28.1\\
DSTC non-parametrized           &   6398K  & 27.6   \\ 
DSTC parametrized              & 6346K    & \textbf{26.3} \\ 
\end{tabular}
}
\caption{Image generation on our implementation of the DCGAN \cite{radford2016}, trained on CelebAScaled and evaluated with Fr\'{e}chet Inception distance (FID) scores (lower is better).  We replace the last transposed convolutional layer $ K=4 $ of the DCGAN generator.  With the Pixel-wise Shuffle, or CARAFE, the DCGAN training did not converge.}\label{DCGAN_Res}
\end{table}
We present quantitative results in table~\ref{hrnet:table} and note the DSTC outperforms the other baselines.  We also present qualitative results in figure~\ref{hrnet_results:fig}.

\subsection{Image Generation with DCGAN}\label{dcgan:sec}

\noindent We use the Deep Convolutional Generative Adversarial Network (DCGAN) \cite{radford2016}, a well-known generative model, to create synthetic faces at different scale.  We use the DCGAN from the \cite{PYTORCH} repository, illustrated in the supplementary materials, which has four upsampling/downsampling TCs of kernel size $ K = 4 $ in the generator/discriminator.
\\
\indent We use celebrity faces \cite{CelebA}, scaled at $ \{\frac{1}{4}, \frac{1}{2}, \frac{3}{4},1\} $, with shape $ 64 \times 64 $ -- the input image size of the original DCGAN, and we split the images into 800K training set, 100 validation/development set and 300 test set.  We use the same training procedure as \cite{radford2016} with the Fr\'{e}chet Inception distance (FID) \cite{heusel2017} for evaluation.  We provide further details on the dataset, training and evaluation in the supplementary materials.  For each experiment we train three models, after each epoch we calculate the FID score between generated images and the validation set.  We pick the best model on these validation scores and evaluate it on the test set.
\\ \indent In our task, we simply replace the last upsampling TC layer of the Generator $ G $.  As the Transposed Pixel-Adaptive Convolution is only implemented for odd $ K $, we evaluated this layer for $ K=3,5 $.  We present results in table \ref{DCGAN_Res}.
Our DSTC layer outperforms all of the baseline layers.  Furthermore, we note that when substituting the pixel-wise shuffle + conv. or CARAFE layer, the adversarial training did not produce recognizable faces.

\subsection{3D Diffusion MRI Enhancement}\label{diqt:sec}

\begin{figure*}[ht]
\begin{center}
   \includegraphics[width=\linewidth]{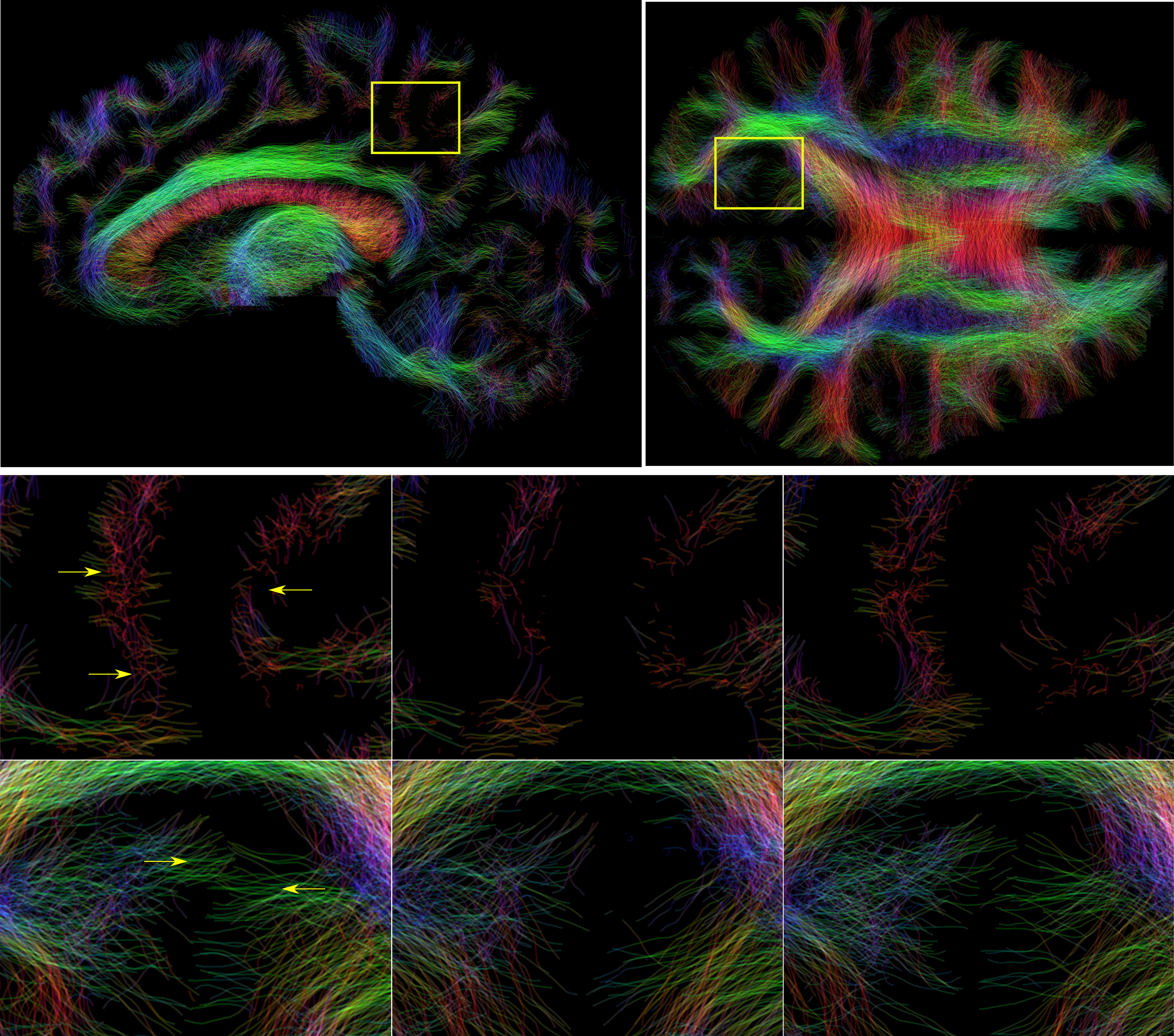} 
\end{center}
   \caption{Probabilistic brain tractography from MAP-MRI coefficients. The streamlines show estimated pathways of brain connections, see e.g. \cite{berg2014} different colors correspond to different streamline direction: red - left to right; green - front to back; blue - top to bottom of the brain.  Top row: Whole brain probabilistic tractography from the image reconstructed from low-resolution of the DIQT w. DSTC.  Bottom two rows:  Zoomed-in regions from i) ground truth (left), ii) baseline DIQT \cite{blumberg2018} (middle), iii) our DIQT with DSTC (right).  The tractography on baseline DIQT misses association fibres in the parietal lobe and the occipital lobe, that the ground truth and DIQT with DSTC finds.}
\label{DIQT_tractography:fig}
\end{figure*}

\begin{table}[ht]
\centering
\resizebox{\linewidth}{!}{
\begin{tabular}{c c c c c c}
\multicolumn{1}{c}{\multirow{2}{*}{\pbox{2cm}{Model}}} & 
\multicolumn{1}{c}{\multirow{2}{*}{\pbox{2cm}{NRL}}} & 
\multicolumn{1}{c}{\multirow{2}{*}{\pbox{2cm}{Params.}}} & 
\multicolumn{3}{c}{Brain Region} \\
\cline{4-6} 
 &  &  & Interior & Exterior & Total  \\ 
\hline
DIQT State-Of-Art \cite{blumberg2018} & 4 & 876K & $ 5.58 \pm 0.25 $ & $ 12.13 \pm 1.24 $ & $ 8.46 \pm 0.67 $ \\
\hline
DIQT w. DSTC & 4 & 888K & $ \textbf{5.24} \pm 0.25 $ & $ \textbf{12.05} \pm 1.27 $ & $ \textbf{8.27} \pm 0.70 $   \\
DIQT w. DSTC & 3 & 705K & $ 5.25 \pm 0.25 $ & $ \textbf{12.05} \pm 1.22 $ & $  \textbf{8.27} \pm 0.67 $   \\
DIQT w. DSTC & 2 & 522K & $ 5.33 \pm 0.25 $ & $ 12.13 \pm 1.27 $ & $ 8.35 \pm 0.69 $ 
\end{tabular}
}
\caption{Root-Mean-Squared-Error (lower is better) between the image reconstructed from low resolution with the original high resolution image of 8 test subjects.  We replace the 3D pixelwise-shuffle of the DIQT with the 3D DSTC and vary the number of reversible layers (NRL) per stack.}\label{DIQT:table}
\end{table}

\noindent Image Quality Transfer (IQT) is a paradigm for propagating information from rare and expensive high-quality acquisitions, to standard, more readily available acquisitions \cite{alexander2017,blumberg2018,lin2019,tanno2021}.  IQT involves downsampling high-quality acquisitions to produce a proxy for a mundane clinical scanner and then using patch-based supervised learning to enhance the image quality of the standard quality images to approximate that of the high quality images.  This technique has been shown to improve both visual image quality and performance in downstream analysis tasks such as brain-connectivity mapping \cite{alexander2017} and epileptic lesion conspicuity in images from low-field scanners in low-and-middle-income countries \cite{lin2019}.  The state-of-the art approach used in IQT for enhancing 3D human-brain diffusion MRI is the Deeper Image Quality Transfer Network (DIQT) \cite{blumberg2018}, which provides the minimum reconstruction errors on a standard test set and also was recently adapted to the related task of harmonizing data across different scanner centers and acquisition protocols \cite{blumberg2019,ning2019}.  As noted earlier, we take the opportunity to reinforce the novel contribution of implementing the 3D DSTC, by investigating whether we can improve the performance of the DIQT with our novel layer.
\\
\indent The DIQT network, illustrated in the supplementary materials, has three 3D convolutional layers followed by a 3D Pixel-wise upsampling shuffle \cite{shi2016}, where each convolutional layer is preceded by $ NRL \in \mathbb{N} $ reversible layers (RLs) \cite{gomez2017}, this formulation allowed the users to integrate a novel low-memory technique, allowing it to manage the high memory demands of applying deep learning to multiple-channeled, high-resolution, medical imaging data.
\\
\indent For direct and fair comparison with the previous state-of-the-art \cite{blumberg2018} we used the same dataset, preprocessing, training procedure, and evaluation as \cite{blumberg2018}, described in detail in the supplementary materials.  We simply replace the sub-pixel convolutional layer in the DIQT with our DSTC layer.   We then reduced the number of reversible layers (NRL) per stack (which had been optimized for performance in \cite{blumberg2018}) and present quantitative results in table~\ref{DIQT:table}, where we obtain state-of-the art results, even with fewer parameters.  We show qualitative results in tractography in figure~\ref{DIQT_tractography:fig} and other qualitative improvements in the supplementary materials.

\section{Conclusion}
\noindent In this paper, we introduced a novel upsampling layer in 2D,3D that improves decoding by handling deformations. We demonstrate performance enhancement in a diverse set of application tasks, with a small number of parameter increase. Our layer can be used as a drop-in replacement for TC and other upsampling operators and the code will be publicly available.

\subsection*{Acknowledgements}

\noindent We greatly thank Tristan Clark, Matteo Figini, Adriano Koshiyama and thank Yipeng Hu, Ed Martin, James O'Connor.  SB is supported by an EPRSC and Microsoft scholarship and EPSRC grants M020533 R006032 R014019, MX by GSK funding (BIDS3000034123) via UCL EPSRC CDT in i4health and UCL Engineering Dean's Prize.  This work was also supported by the NIHR ULCH Biomedical Research Centre.

{\small
\bibliographystyle{ieee_fullname}
\bibliography{main}
}

\newpage

\section*{Supplementary Materials}

\section*{Additional Experimental Details}

\noindent \textbf{Mask-RCNN Additional Details} We present more details of our settings in section~\ref{mrcnn_fpn:sec}, where we used the Mask-RCNN, which extends the Faster-RCNN \cite{ren2015}, which introduced Region Proposal Networks (RPN), using CNNs to propose regions, that were then passed to a classifier in the final stage of object detection.  We used the standard 1x schedule from \cite{mmdetection}.  During training, the images were resized to shape $ 1333 \times 800 $ and flipped with probability $ 0.5 $.  We trained the networks for 12 epochs, batch size 16 with SGD optimizer with momentum 0.9.  There were 500 warm-up iterations and during training the learning rate started at 0.02, dropping by a factor of 10 at epochs 8,11.  We used multi-scale testing and uploaded the predictions to the server to obtain the results on the latest (2019) test-dev set.
\\
\\ \noindent \textbf{HRNet Additional Details} We describe further details of our settings used in section~\ref{hrnet:sec}. We use the standard 20K schedule from \cite{mmseg2020}.  With a batch size of 16 across 4 or 8 GPUs we train for 20K iterations with SGD optimizer, weight decay $ 0.0005 $ and learning rate decaying polynomially from $ 0.01 $ to $ 0.0001 $.  During training the images are rescaled to $ 2048 \times 512 $, cropped to $ 512 \times 512 $ and randomly flipped with probability $ \frac{1}{2}$.
\\
\\ \noindent \textbf{DCGAN Additional Experimental Details} We provide more details of our settings in section~\ref{dcgan:sec} where we used the DCGAN, that consists of a set of constraints on the topology of convolutional GANs, that improve training stability and are shown to learn a good hierarchy of representations from object parts to scenes.  
\\ \indent Our dataset is scaled faces from real celebrities, which we denote as CelebAScaled.  We first crop high-quality images of celebrities from \cite{CelebA}, to a $ 64 \times 64 $ region around the subject's face, which is the input size of the original DCGAN implementation.  We then cropped-and-rescaled each image with scaling factors $ \{\frac{1}{4}, \frac{1}{2}, \frac{3}{4},1\} $ quadrupling the size of CelebA.  We split the images into 800K training set, 100 validation/development set and 300 test set.
\\ \indent During training, where we first draw $ x \in \mathcal{N}(0,1)^{100} $, the generator produces a fake image $ \widetilde{y} = G(x) \in \mathbb{R}^{3 \times 64 \cdot 64} $.  The discriminator $ D : \mathbb{R}^{3 \times 64 \cdot 64} \rightarrow [0,1] $ attempts to classify both real images $ y $ and fake images $ \widetilde{y} $ correctly.  Both networks are trained in an adversarial fashion with batch size $ 128 $, ADAM optimizer with betas $ (0.5,0.999) $ and learning rate $ 0.0002 $.
\\
\indent To evaluate the generated images, we use the Fr\'{e}chet Inception distance (FID) \cite{heusel2017}, which compares two sets of images from different distributions, and has been used in recent GAN papers \cite{park2019}.  This metric compares the similarity of two sets of images, via a similarity measure of intermediate feature maps, when the images are passed through a pre-trained network.  This is defined as follows.  Suppose we have two sets of images and the InceptionV3 network \cite{szegedy2016}, pre-trained on ImageNet \cite{IMAGENET}.  We calculate respective means $ \mu_1,\mu_2 $ and covariances $ \sigma_1, \sigma_2 $, of the 2048-dimensional activations of the InceptionV3 pool3 layer.  The FID score is $ FID \ = \ || \mu_1 -\mu_2 || + Trace(\sigma_1 + \sigma_2 - 2(\sigma_1 \sigma2)^{\frac{1}{2}}) $, where lower scores signifies that the two sets of images are more similar to each other.
\\
\\ \noindent \textbf{DIQT Additional Details} We provide further details for our settings in section~\ref{diqt:sec}, where for direct and fair comparison with the previous state-of-the-art we used the same dataset, preprocessing, training procedure, and evaluation as \cite{blumberg2018}.  We used 40 brain scans of healthy young adults from the Human Connectome Project \cite{sotiropoulos2013}.  Each scan consisted of 90 diffusion weighted images with voxel size $ 1.25mm^{3} $ total volume $ 145 \times 174 \times 145 $, of which $ 29\% $ is brain tissue.  Then we extracted the diffusion tensor images (DTI), measuring water diffusivity, producing $ 6 $ channels per voxel; and the MAP-MRI coefficients \cite{ozarslan2013} which generalizes DTI producing several novel parameters to capture previously obscured microstructural features, for the 16 scans in \cite{alexander2017}.  The low-resolution images, a proxy for acquisitions obtained from normal scanners, were obtained by downsampling these images.  We used 32 subjects for training and 8 for testing for table~\ref{DIQT:table}, where the root-mean-squared-error (RMSE) on brain tissue only, is used for evaluation.
\\ 
\indent We used identical training procedure and training hyperparameters as \cite{blumberg2018}, to make a fair comparison with \cite{blumberg2018}.  As entire brain volumes are too large for end-to-end deep learning training, we performed our training patch-wise where  patches of input/target shape are $ 11^{3},14^{3} $, with the patch center voxel within the brain tissue.  We separated the patches from the training subjects ($ \approx 72K $ patches) into 80\%-20\% training-development set.  We used the ADAM optimizer, with learning rate $ 0.001 $, batch size $ 12 $ and MSE loss.  When predicting on the test subjects, we parcellated the low-resolution image into patches and concatenated the target patch predictions.  We trained four models per experiment and then evaluated the best performing model on the validation set, on the test set.

\clearpage
\newpage
 
\section*{Ablation Study for Interpolation Kernel Hyperparameters}

\begin{figure}[ht]
    \centering   
    \resizebox{\linewidth}{!}{
    \includegraphics[scale=0.4]{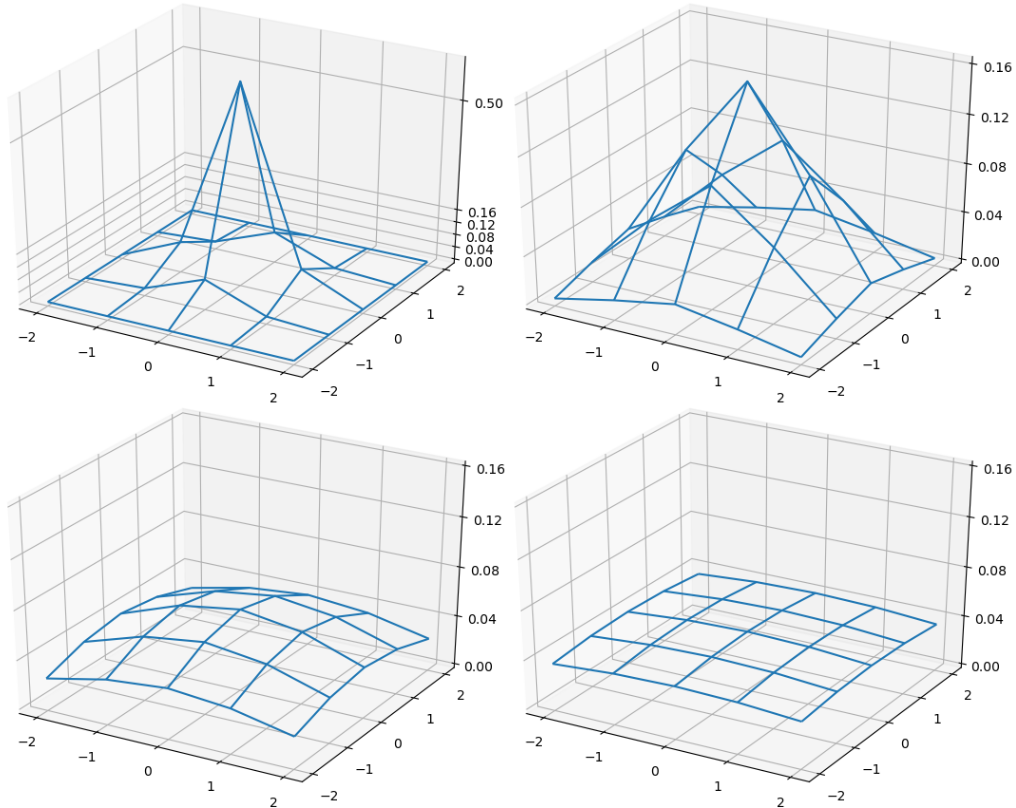}
    }
   \caption{2D Gaussian plots to choose Gaussian variances $ \Sigma_{i} $ for kernel size $ K_{\Sigma} = 5 $, for our interpolation kernel $ G $ in equation~\ref{InterpEq}.  We interpolate value $ 1 $ at location $ p = (0,0) $ to locations on the 2D grid $ q = \{-2,-1,0,1,2\}^{2} $, and plot value $ G(p,q) $.  Variances are $ 2^{-2} $ top-left, $ 2^{0} $ top-right, $ 2^{2} $ bottom-left, $ 2^{4} $ bottom-right.
   }\label{Gaussian_plots:fig}
\end{figure}

\noindent We performed a brief hyperparameter search to pick the hyperparameters for our interpolation kernel, defined in equation~\ref{GaussInt}.  This includes the number of Gaussians in our interpolation kernel ($ s $), the variances for these Gaussians ( $ \Sigma_{0} < \Sigma_{1} < ... $ ) and the side of the interpolation kernel ($ K_{\Sigma} $).  We considered four  different Gaussian variances $ \Sigma_{0,1,2,3}=\{2^{-2},2^{0},2^{2},2^{4}\} $ (note the standard deviations are $\frac{1}{2},1,2,4 $), chosen due to their different spreads, which may be seen visually in figure~\ref{Gaussian_plots:fig}.  We also performed a brief ablation study with the experimental settings in section~\ref{mrcnn_fpn:sec}, and report results for different combinations of of Gaussian variances in table~\ref{interpolation_kernel_variances:table}.  We performed an additional ablation study in the same experimental settings, to investigate the size of the interpolation kernel in table~\ref{interpolation_kernel_size:table}.

\begin{table}[ht]
\centering
\resizebox{\linewidth}{!}{
\begin{tabular}{c c c c c }
\multicolumn{2}{c}{\multirow{1}{*}{Interpolation Kernel}} & &
\multicolumn{2}{c}{\centering COCO Val} \\
\cline{1-2} \cline{4-5}
\multicolumn{1}{c}{\multirow{1}{*}{Type}} & \multicolumn{1}{c}{\multirow{1}{*}{\pbox{1.5cm}{\centering Variances$ \ \Sigma_{i} $}}} & & \multicolumn{1}{c}{\multirow{1}{*}{\pbox{1.5cm}{\centering Box AP}}}  & \multicolumn{1}{c}{\multirow{1}{*}{\pbox{1.5cm}{\centering Mask AP}}} \\ 
\hline
Bilinear &  -- & & 38.6 & 35.1 \\ 
\hline
Ours  & \{0.25\} & & 38.7  &  35.3 \\ 
Ours  & \{1\} & & 38.6 & 35.1 \\ 
Ours  & \{4\} & & 38.9 & 35.4 \\ 
Ours  &  \{16\} & & 38.7 & 35.2 \\ 
Ours  &  \{0.25,1,4,16\} & & 38.9 & 35.6 \\ 
\end{tabular}
}
\caption{Interpolation kernel hyperparameter search / ablation study, for the interpolation kernel Gaussian variances $ \Sigma_{i} $, experimental settings from section~\ref{mrcnn_fpn:sec}.  We considered variances $ \Sigma_{0,1,2,3}=\{2^{-2},2^{0},2^{2},2^{4}\} $ by picking four variances with different spreads, see e.g. figure~\ref{Gaussian_plots:fig}. }\label{interpolation_kernel_variances:table}
\end{table}

\begin{table}[ht]
\centering
\resizebox{\linewidth}{!}{
\begin{tabular}{c c c c c }
\multicolumn{2}{c}{\multirow{1}{*}{Interpolation Kernel}} & &
\multicolumn{2}{c}{\centering COCO Val} \\
\cline{1-2} \cline{4-5}
\multicolumn{1}{c}{\multirow{1}{*}{Type}} & \multicolumn{1}{c}{\multirow{1}{*}{\pbox{1.5cm}{\centering Size$ \ K_{\Sigma} $}}} & & \multicolumn{1}{c}{\multirow{1}{*}{\pbox{1.5cm}{\centering Box AP}}}  & \multicolumn{1}{c}{\multirow{1}{*}{\pbox{1.5cm}{\centering Mask AP}}} \\ 
\hline
Bilinear &  2 & &  38.6 & 35.1 \\ 
\hline
Ours  &  3  & & 38.8  & 35.6 \\ 
Ours  &  5  &  & 39.0 & 35.5 \\ 
Ours  &  7  &  & 38.9 & 35.6  \\ 
Ours  &  9  &  & 38.8 & 35.4   \\ 
\end{tabular}
}
\caption{Interpolation kernel hyperparameter search / ablation study, for the interpolation kernel size $ K_{\Sigma} $, experimental settings from section~\ref{mrcnn_fpn:sec}.  We used $ K_{\Sigma} = 5 $ in the main paper.}\label{interpolation_kernel_size:table}
\end{table}

\clearpage
\newpage

\section*{Code Usage}

\noindent Below we illustrate the standard usage of the 2D TC from the PyTorch \cite{PYTORCH} library:

\begin{lstlisting}[language=Python, basicstyle=\footnotesize,tabsize=2]

from torch.nn import ConvTransposed2d
layer = ConvTransposed2d(
  in_channels,
  out_channels,
  kernel_size,
  stride,
  padding,
  output_padding,
  groups,
  bias,
  dilation,
  padding_mode,
)
\end{lstlisting}

\noindent Our DSTC layer is also implemented as a PyTorch layer and is available in 2D or 3D.  The DSTC takes in analogous arguments to the prototype ConvTransposed2d/ConvTransposed3d layer, in addition to additional arguments that correspond to our modules:

\begin{lstlisting}[
language=python,
basicstyle=\footnotesize,
escapeinside={`}{'},
keepspaces=true,
]

import DeformablyScaledConvTranspose as DSTC
layer = DSTC(
  in_channels,
  out_channels,
  kernel_size,
  stride,
  padding,
  output_padding,
  groups,
  bias,
  dilation,
  padding_mode,
  dimension, # {2,3}
  offset_version, # {off,unparametrized,parametrized}
  interpolation_kernel, # {bilinear, gaussian}
  Gaussian_variances, # Gaussian variances `$\Sigma_{i}$'
  Gaussian_version, # {unparametrized, parametrized}
  K_Sigma, # Interp. Kernel Size `$K_{\Sigma}$'
  module_channels, # Chans before/after compression
  skip, # Skip connection between `$X,Y$'
  scatter_loop, # trade memory w. computational time
)

\end{lstlisting}
Our code will be publicly available.

\newpage

\section*{Additional Visualizations}

\begin{figure}[ht]
\resizebox{\linewidth}{!}{
    \centering   
    \includegraphics[width=\linewidth]{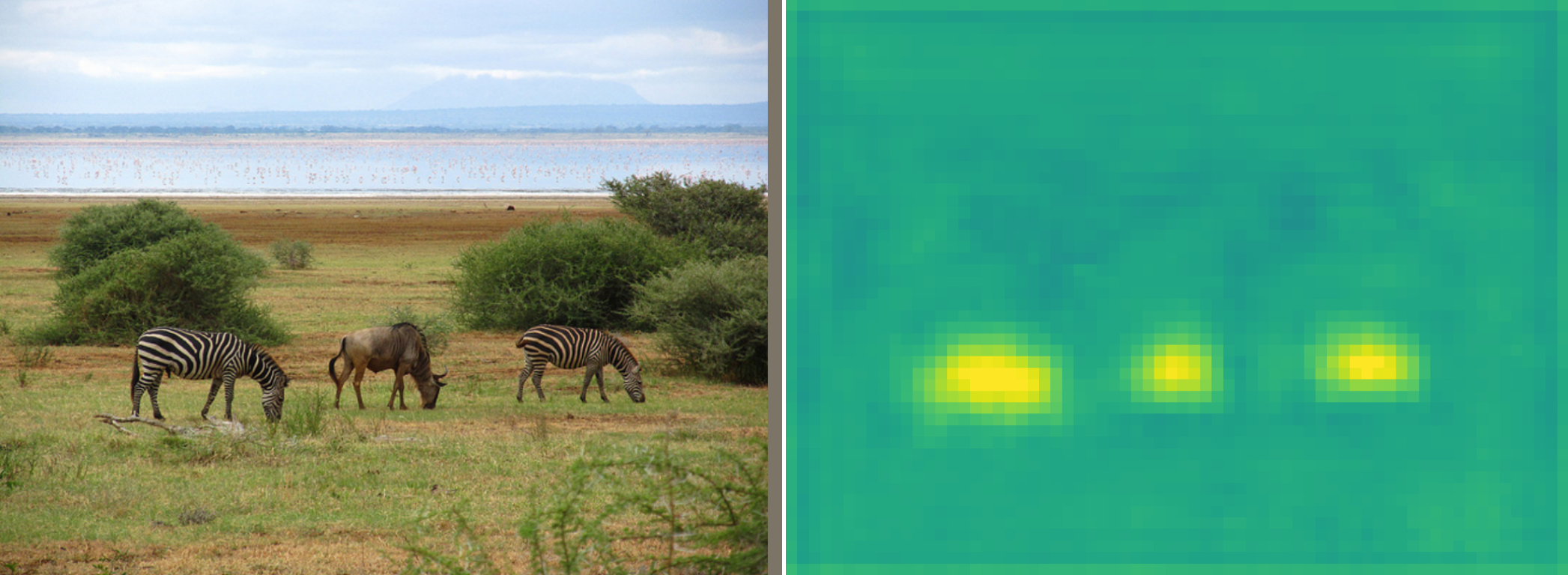}
    }
   \caption{LHS: Input image, RHS: Dilation learnt by the DSTC.  Experimental settings from section~\ref{mrcnn_fpn:sec}}
\end{figure}

\begin{figure}[ht]
\resizebox{\linewidth}{!}{
    \centering   
    \includegraphics[width=\linewidth]{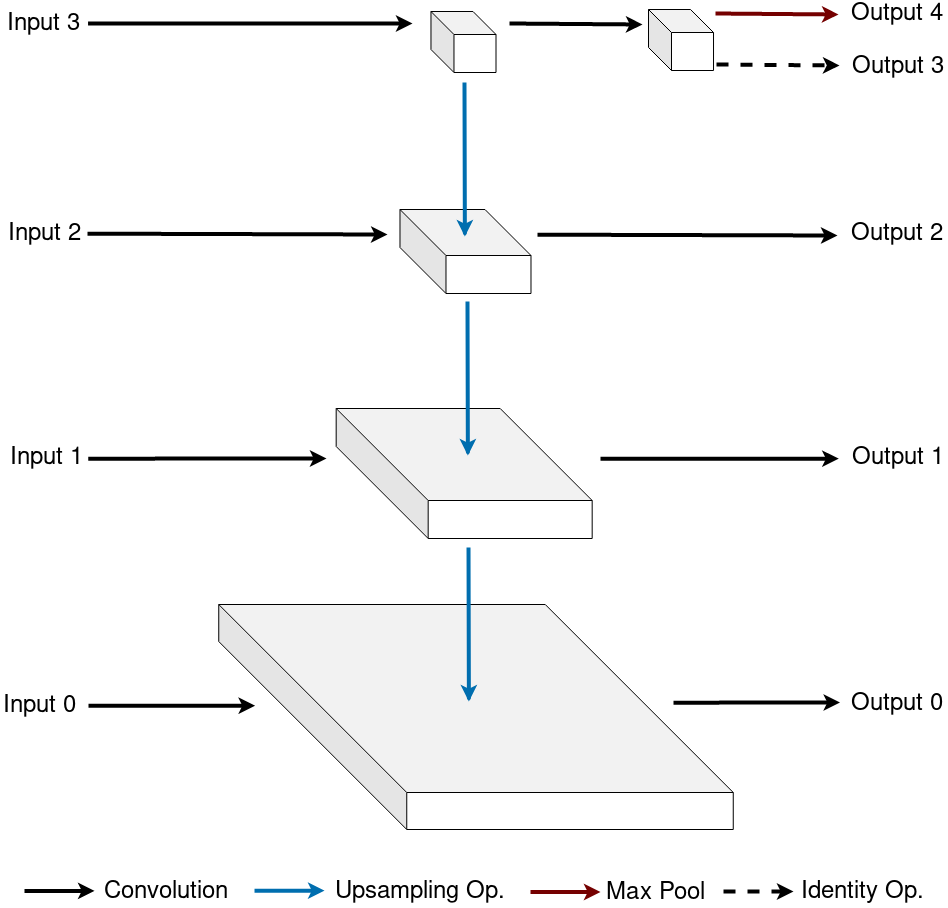}
    }
    \caption{The Feature Pyramid Network (FPN) \cite{lin2017} that we use in section~\ref{mrcnn_fpn:sec}.  We replace the upsampling operators in the top-down pathway (three blue lines).  Feature maps are summed.}
\end{figure}

\begin{figure*}[ht]
\begin{center}
    \includegraphics[width=0.9\linewidth]{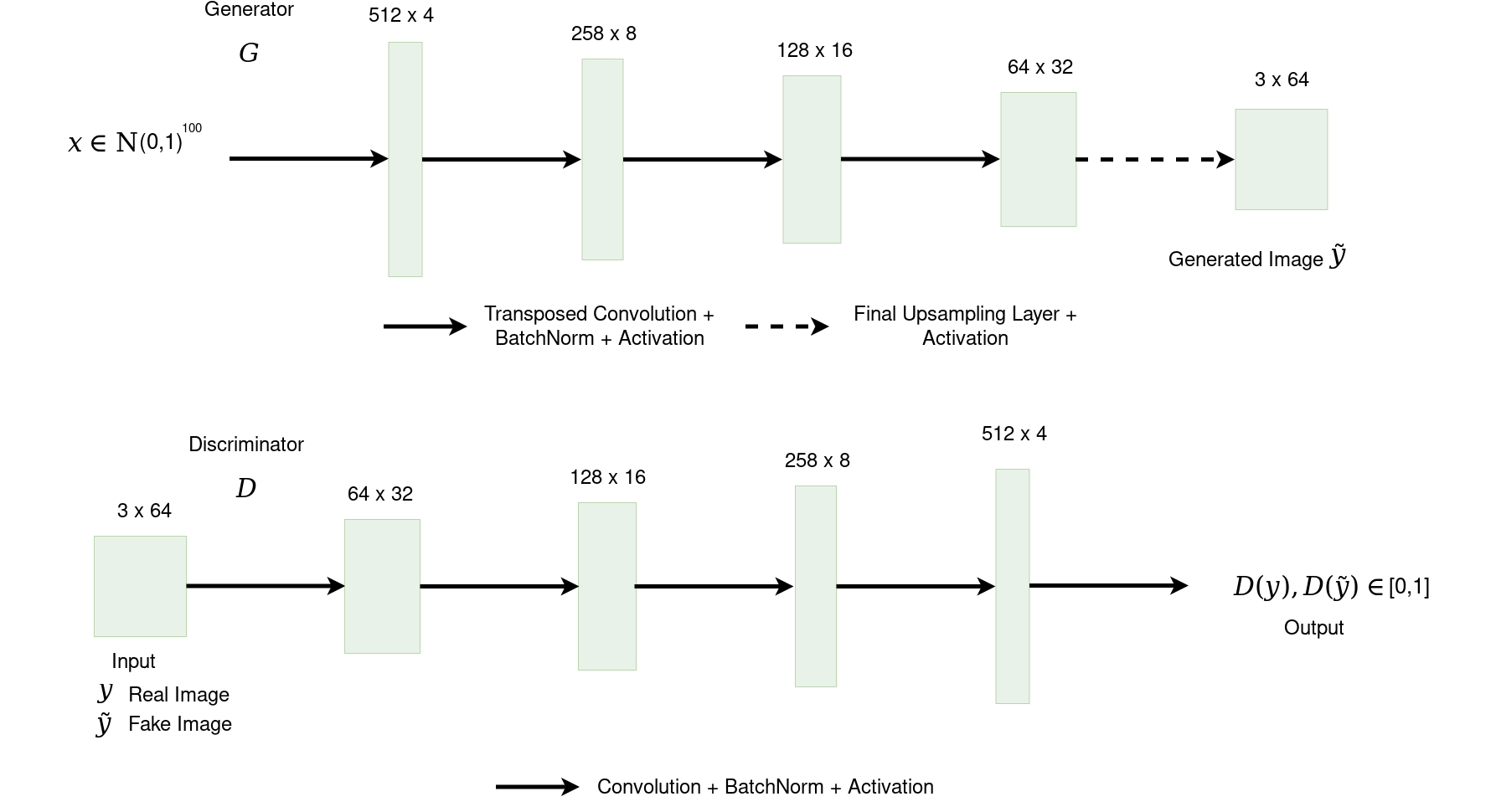}
    \caption{The DCGAN \cite{radford2016} that we used in section~\ref{dcgan:sec}  In the main text we replace the last TC in the generator (dotted line) with various upsampling operators.}
\end{center}
\end{figure*}

\begin{figure*}[ht]
\begin{center}
    \includegraphics[width=0.9\linewidth]{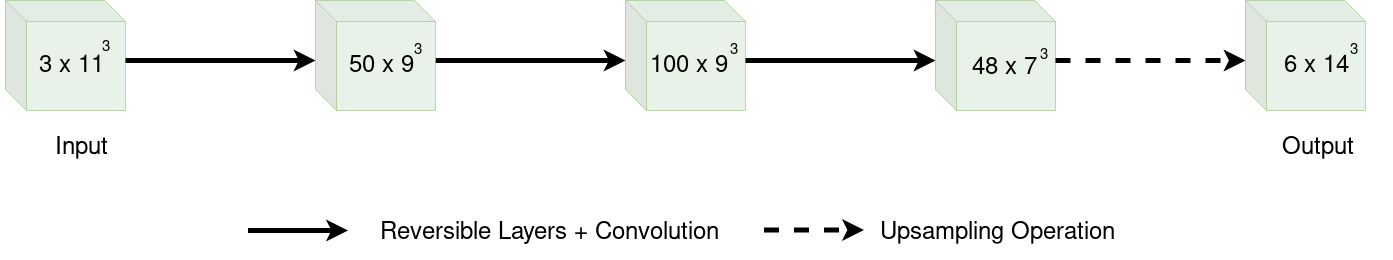}
    \caption{The DIQT \cite{blumberg2018} that we introduced in section~\ref{diqt:sec}.  Each thick black line consists of $ NRL $ number of reversible blocks \cite{gomez2017}, stacked in succession, followed by a convolution.  The upsampling operation is the pixelwise shuffle \cite{shi2016}, replaced in the main text.  The network takes a $ 11^{3} $ spatial patch in low-dimension space and predicts a $ 14^{3} $ patch in high-dimensional space (corresponding to $ 7^{3} $ in low-dimensional space).}
\end{center}
\end{figure*}

\begin{figure*}[ht]
    \centering
    \includegraphics[scale=0.4]{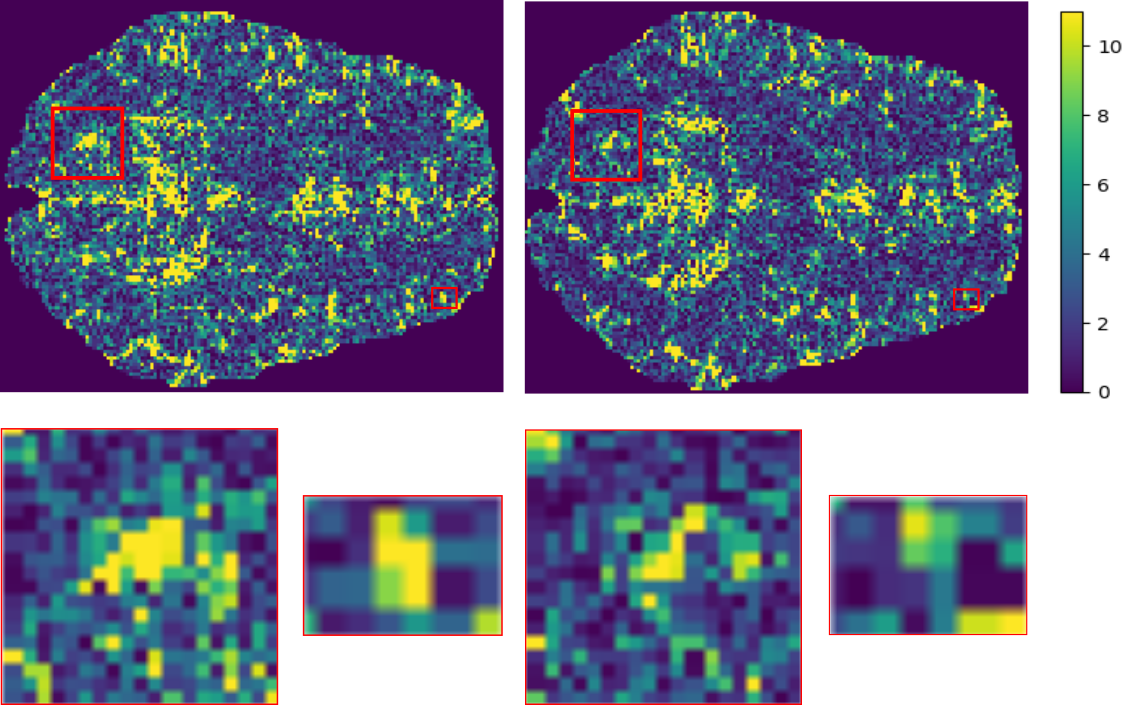}
    \includegraphics[scale=0.4]{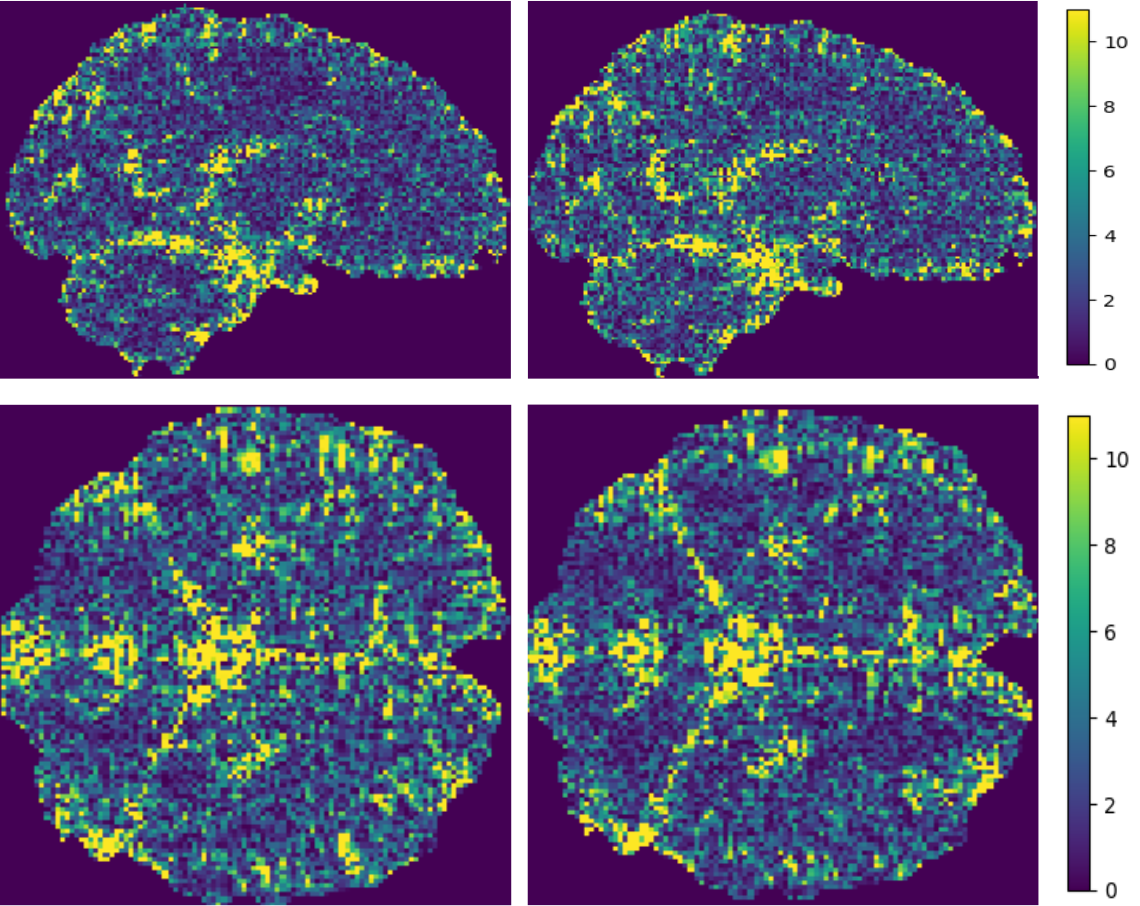}
    \caption{Mean-Squared-Error (yellow is high) of a 2D slice from a 3D prediction of a test subject,  LHS: state-of-the-art DIQT, RHS: we replace the final PS layer with the DSTC.}\label{DIQT_MSE:im}
\end{figure*}

\end{document}